\documentclass[11pt]{article}

\usepackage[margin=1.25in]{geometry}
\usepackage{tablefootnote}

\usepackage{comment}
\usepackage{times}

\usepackage{hyperref}       
\usepackage{url}
\usepackage{url}            
\usepackage{booktabs}       
\usepackage{amsfonts}       
\usepackage{nicefrac}       
\usepackage{microtype}      

\usepackage{booktabs}
\usepackage{subfigure}
\usepackage{bm}
 \usepackage{algorithm}
 \usepackage{algorithmic}
\usepackage{amssymb}
\usepackage{amsmath}
\usepackage{amsfonts}
\usepackage{paralist}
\usepackage{comment}

\usepackage{amsmath,amsfonts,amsthm,amssymb}
\usepackage{graphicx}
\usepackage{wrapfig}
\usepackage{lipsum}
\usepackage{url}
\usepackage{multirow}
\usepackage{authblk}

\def \x {\mathbf{x}}
\def \c {\tilde c}

\def \c {\mathbf{c}}

\def \u {\mathbf{u}}
\def \v {\mathbf{v}}
\def \tv {\tilde{\mathbf{v}}}

\def \x {\mathbf{x}}

\def \R {\mathbb{R}}

\def \a {\mathbf{a}}

\def \b {\mathbf{b}}

\def \x {\mathbf{x}}
\def \c {\tilde c}

\def \c {\mathbf{c}}

\def \B {\mathbf{B}}
\def \X {\mathbf{X}}
\def \Y {\mathbf{Y}}
\def \bmY{\bm{\mathcal{Y}}}

\def \U {\mathbf{U}}
\def \V {\mathbf{V}}

\def \u {\mathbf{u}}
\def \v {\mathbf{v}}

\def \x {\mathbf{x}}

\def \R {\mathbb{R}}

\def \bmX {\bm{\mathcal X}}

\def \supp {\mathrm{supp}}

\title{Negative-Unlabeled Tensor Factorization for Location Category Inference from Highly Inaccurate Mobility Data}
\author[1]{Jinfeng Yi}
\author[2]{\  Qi Lei\footnote{Part of the work was done during Qi Lei's internship at IBM Research.}}
\author[1]{\  Wesley Gifford}
\author[3]{\  Ji Liu}
\author[1]{\  Junchi Yan}
\affil[1]{IBM Research}
\affil[2]{University of Texas at Austin}
\affil[3]{University of Rochester}

\begin{document}
\maketitle
\begin{abstract}
Identifying significant location categories visited by mobile users is the key to a variety of applications. 
This is an extremely challenging task due to the possible deviation between the estimated location coordinate and the actual location, which could be on the order of kilometers. 
We consider the location uncertainty circle determined by the reported location coordinate as the center and the associated location error as the radius. Such a location uncertainty circle is likely to cover multiple location categories, especially in densely populated areas.  
To estimate the actual location category more precisely, we propose a novel tensor factorization framework, through several key observations including the intrinsic correlations between users, to infer the most likely location categories within the location uncertainty circle. In addition, the proposed algorithm can also predict where users are even in the absence of location information. In order to efficiently solve the proposed framework, we propose a parameter-free and scalable optimization algorithm by effectively exploring the sparse and low-rank structure of the tensor. Our empirical studies show that the proposed algorithm is both efficient and effective: it can solve problems with millions of users and billions of location updates, and also provides superior prediction accuracies on real-world location updates and check-in data sets.
\end{abstract}

%
%



\maketitle

\section{Introduction}\label{sec:intro}
Understanding mobile users' spatio-temporal activities is a central theme in a variety of applications, including personalized
advertising \cite{bao2015recommendations}, 
traffic monitoring \cite{mohan2008nericell}, security management \cite{zheng2014urban}, and assistance of the elderly and disabled \cite{barbeau2010travel}. To this end, a key step is to identify the significant location categories, such as restaurants, gyms, and shopping malls, visited by each mobile user from her mobile location updates. Unfortunately, this is an extremely challenging task since mobile location updates are often highly inaccurate for a number of reasons, for example,
\begin{itemize}
\item {\bf Signal condition and quality}: In GPS-based systems signal conditions, such as dense foliage or urban canyons, can impact the ability to communicate with the required number of satellites or introduce delay in signal propagation -- ultimately leading to decreased accuracy.
\item {\bf Bias of location estimation techniques}: For triang-ulation-based methods, the density of WiFi or cellular networks can significantly affect accuracy. The fact that the density of such networks varies significantly across locations means different levels of accuracy will be observed.
\item {\bf Device limitations}: Many mobile applications may deliberately request less frequent or less accurate location information in order to conserve battery power for applications where high accuracy is not explicitly required.
\end{itemize}

\begin{figure}[t]
\centering
\includegraphics[width=0.8\columnwidth]{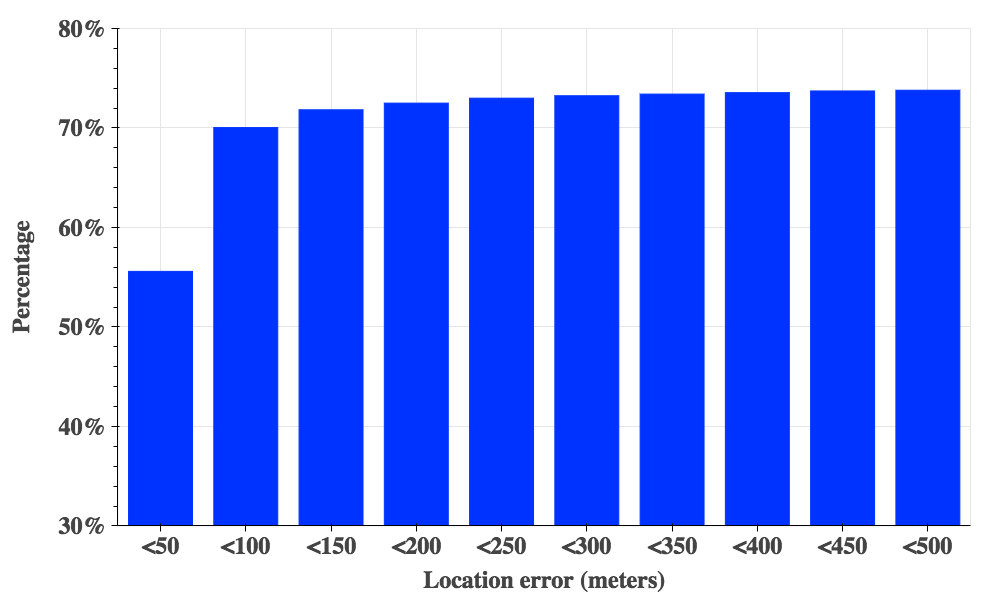}
\caption{Histogram of mobile phone location errors for a large anonymized sample of mobile location data.}\label{fig1}
\end{figure}
\begin{figure}[t]
\centering
\includegraphics[width=0.8\columnwidth]{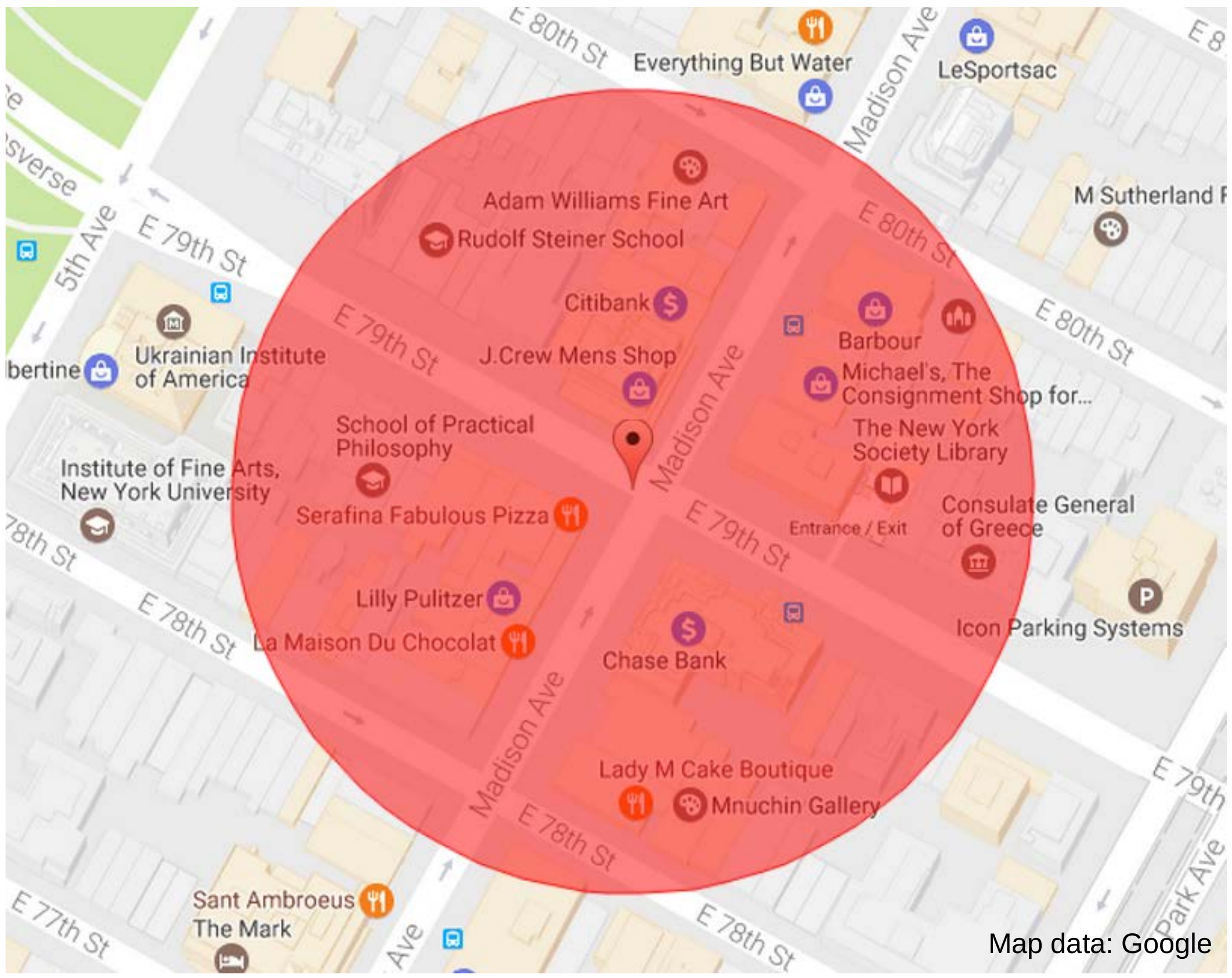}
\caption{An example of location uncertainty circle with a 100-meter radius in New York City. It covers 
seven location categories: \emph{library, restaurant, school, shop, gallery, bank,} and \emph{consulate}.}\label{fig2}
\end{figure}

Using a large anonymized sample of mobile location data
collected as part of our partnership with a leading U.S.\ location-based services company, we investigated the distribution of location errors. 
The error distribution collected from a major U.S.\ city is depicted in Figure~\ref{fig1}. The data shows that only $56\%$ of updates have location errors within $50$ meters, while $26\%$ of the updates have location errors greater than $500$ meters. Using the estimated location coordinate as the center and its associated error as the radius, we can draw a circle where the mobile user may be located. We refer to such a circle as \textit{location uncertainty circle}, which may cover multiple location categories, especially in densely populated areas like cities. Figure~\ref{fig2} shows a location uncertainty circle drawn based on a  simulated location update 
and a $100$ meter location error in New York City. Although this circle is not very large, it still covers seven location categories, including library, restaurant, school, shop, gallery, bank, and consulate, all of which might be the true venue visited by this user. 
To decide a unique location category, a naive idea is to use $1$-nearest neighbor (1-NN) approach, i.e., simply choose the venue that is closest to the estimated location coordinate. However, this approach is problematic since there may exist multiple nearby venues with almost identical distances to the estimated location coordinate. Figure~\ref{fig2} gives one such example. The estimated location coordinate is almost equally close to the boundaries of four venues, including a Chase bank, a pizza restaurant, a clothing shop, and the New York Society Library. Therefore, it is almost impossible to identify the true venue using the 1-NN approach. Besides, since many location updates have very large location errors, and the estimated location coordinate could be highly biased, there is no guarantee that the true venue visited by the mobile user is truly close to the estimated location coordinate. In addition to large location errors, another challenge arises from anonymity: since mobile users are anonymous and we have no access to their personal information, no user feature or side information is available to help locate them, which further increases the difficulty. For this same reason, conventional supervised learning approaches such as feature-based classification cannot be applied to locate mobile users. Given all these limitations, we aim to examine the following challenging question in our study:
\vspace{0.1cm}

\emph{Is it possible to accurately infer mobile users' visited location categories \emph{purely} based on their \emph{highly inaccurate} mobile location updates?}
\vspace{0.1cm}


In this paper, we provide an affirmative answer to this question. Specifically, we develop a novel framework for location category inference based on some key observations. 
We first observe that for each location update, there is one and only one true location category since a user can only visit one place at a time. In addition, since the user's true location must be within the location uncertainty circle, the location categories not covered by this circle cannot be visited. Therefore, the probabilities of visiting these location categories is zero, while the probabilities of location categories covered by the location uncertainty circle sum up to one. Indeed, we can treat this problem as a two-class learning problem where the positive class contains only one example (i.e., the true location category) while all the other examples belong to the negative class (i.e., false location categories). Under this scenario, all the labeled examples are sampled from the negative class while the unlabeled examples come from both negative and positive classes. We refer to such a problem as a \emph{negative-unlabeled} (\emph{NU}) \emph{learning} problem, a counterpart to the positive-unlabeled (PU) learning problem~\cite{DBLP:conf/kdd/ElkanN08,DBLP:conf/icdm/LiuDLLY03,DBLP:conf/icml/HsiehND15} or the problem of learning from implicit feedback in the recommendation literature~\cite{hu2008collaborative,DBLP:conf/uai/RendleFGS09,baltrunas2009towards}. The key feature to PU learning problem is that the labeled examples are only drawn from the positive class, and the unlabeled examples are a mixture of positive and negative class samples.

The observations specified above are insufficient to recover the probabilities of all the location categories since we can assign the non-zero probabilities with any non-negative values that sum up to one. To address this issue, we develop a collaborative approach that exploit all users' location updates to collaboratively locate each other. This is due to the reason that people with the same lifestyle tend to behave similarly. For instance, parents of kindergarteners may drop-off and pick-up their children at roughly the same local time, even if their children are not enrolled in a same kindergarten. Likewise, people with routine work-home schedules tend to go to work from home and go back home after work at similar time, which are the main reasons of morning and evening traffic peaks. In addition to user-user similarity, time-time similarity is another factor we can take advantage of since users' routines are usually consistent from day to day, or from week to week. This observation suggests that the underlying three-way tensor ${\bm{\mathcal X}},$ where $\bm{\mathcal X}_{ijk}$ indicates the probability that user $i$ visited the location category $k$ during the time slot $j$, should be close to low-rank. This observation is also verified by a widely held assumption that people's behaviors are dictated by a small number of latent factors~\cite{wang2014travel,zheng2015trajectory}.

Given the above observations, we first formulate our learning problem as a low-rank tensor factorization problem under NU constraints. However, the high computational cost significantly limits its application to real-world location category inference problem that involves a large number of mobile users and location updates. To address this limitation, we first relax the problem to a matrix optimization problem by converting two three-way tensors to their mode-1 matricizations. We note that the problem after relaxation is still challenging to solve since it requires computation of a low-rank approximation of a large matrix, which is computationally expensive in general. More severely, the optimization problem involves ($\#\text{users}\times \#\text{time slots} \times \#\text{location categories}$) entries, thus a naive optimization algorithm will take at least $O(\#\text{users}\times \#\text{time slots} \times \#\text{location categories})$ time to optimize the problem. To overcome this limitation, we develop an efficient alternating minimization algorithm by effectively exploring the sparsity of the large matrix. Our experiments show that our algorithm is extremely efficient and can solve problems with millions of users and billions of location updates. For instance, our synthetic study shows that the proposed algorithm is able to \textit{perfectly} predict the underlying location categories for $3$ million mobile users and more than a billion location updates in less than $1$ hour.

Finally, we emphasize the key reasons why we specifically focus on the problem of inferring the true location category rather than the actual venue. Firstly, the number of venues is substantially larger than the number of categories. Since the number of users having venue-wise similar behavioral patterns is substantially lower than the number of users having category-wise similar behavioral patterns, it becomes very difficult to leverage collaborative techniques if individual venues are sought. More importantly, knowledge of the categories visited by users often provides enough useful information for many important tasks, such as personalized advertising, customer profiling, and urban planning.

In summary, our main contributions are threefold:
\begin{itemize}
\item To the best of our knowledge, this is the first work that can infer users' location categories purely based on highly inaccurate mobility data. Specifically, we propose a novel learning framework that can not only infer location categories in a collaborative manner, but also handle the issues when large number of users' location updates are sparse and noisy.
\item To the best of our knowledge, this is the first work that studies the tensor factorization problem under the negative-unlabeled (NU) constraints. One advantage of casting the location category inference problem as a tensor factorization problem is that by completing the tensor, we can even infer users' location categories when there is no location update data available.
\item By effectively exploiting sparsity, we develop an extremely efficient algorithm that is able to infer location categories with a huge number of mobile users and location updates.
\end{itemize}

\section{Related Work}\label{sec:background}
To the best of our knowledge, this paper makes the first attempt to learn mobile users' location categories from highly inaccurate mobility location data. We review two existing works related to our study: stay point detection and location semantic meaning identification.

\vspace{0.2cm}
{\noindent \bf Stay point detection}\ \
In a trajectory or sequence of location updates, stay points are defined as the important locations where people have stayed for a while~\cite{zheng2015trajectory}.~\cite{li2008mining} proposed the first stay point detection algorithm that checks if the distance and the time span between an anchor point and its successors in a trajectory are larger than two individual thresholds. If both answers are yes, a stay point is detected. The authors in \cite{yuan2011find,yuan2013t} further improved this algorithm by considering the density of location points. In~\cite{cao2010mining}, stay points were detected by modeling location-location and location-user relationships via a graph-based approach.

\vspace{0.2cm}
{\noindent \bf Location semantic meaning identification}\ \
To go one step further, many location-aware applications also care about the semantic meanings of stay points. To address this problem, a typical idea is to first cluster the stay points to identify regions of interest, and then use a cluster ID to represent stay points belonging to this cluster. Popular clustering approaches in this area include time-based clustering, density-based clustering, and partitioning clustering, as summarized in~\cite{zhou2007mining}. In particular, the authors in \cite{ashbrook2003using} use a variant of $k$-means algorithm to cluster GPS data for detecting users' significant locations. In addition, a density-based clustering algorithm was applied in~\cite{ye2009mining} to infer individual life patterns from GPS trajectory data. The authors in~\cite{xiao2010finding} estimate user similarities in terms of semantic location history using a hierarchical clustering-based approach. The work in \cite{liu2016unsupervised} identifies home and work locations by first transforming user trajectory records into user-location signatures, and then applying $k$-Means clustering on these signatures.

\vspace{0.2cm}
{\noindent \bf Key differences from our work}\ \
Although the problems discussed above share some similarities with our work, they differ from the focus of our paper in the following respects:
\begin{itemize}
\item They usually use accurate location data such as GPS signals to generate trajectories and ignore the location errors. On the contrary, the real-world mobile location update data considered in our study is highly inaccurate with unignorable location errors, making our problem much more difficult to solve.
\item Given a location update in a trajectory, they aim to assign it to a cluster of similar location updates, under the assumption that similar location updates should belong to the same cluster. In comparison, our main focus is to infer the underlying true location category, under a more realistic scenario that even if two location updates appear to be similar, they may belong to different location categories due to unignorable location errors.
\item Most studies mentioned above adopt non-collaborative approaches, where user's profile data is used in isolation to determine her stay points. In contrast, this paper infers the user's location categories together from a unified model that collaboratively locates each user. Comparing with non-collaborative approaches, our approach is more robust to inaccurate and sparse location update data.
\end{itemize}


\section{Methodology} \label{sec:Method}
In this section, we first briefly discuss how to clean the raw mobility location update data and generate the candidate location categories. We then present our negative-unlabeled tensor factorization model, followed by a scalable optimization algorithm.

Throughout the paper, we use boldface Euler script letters, boldface capital letters, and boldface lower-case letters to denote tensors (e.g., ${\bm{\mathcal X}}$), matrices (e.g., $\X$) and vectors (e.g., $\x$), respectively. 
The $(i,j,k)$-th entry of a third-order tensor ${\bm{\mathcal X}}$ and the $(i,j)$-th entry of a matrix ${\X}$ is denoted by $\bm{\mathcal X}_{ijk}$ and $\X_{ij}$, respectively. $|\Omega|$ returns the number of elements in the set of $\Omega$. $[N]$ denotes the set $\{1,2,\cdots, N\}$ for short, where $N$ must be an integral number.

\subsection{Data Preprocessing and Location Candidate Generation} 
Given a collection of mobile location updates in the form \{\emph{anonymous user id, UTC timestamp, estimated location coordinates, location error}\}, we first 
preprocess the raw data. The preprocessing encompasses three key steps:
1) filtering to retain only meaningful location updates, 2) association with location updates with category information, and 3) quantization of time into time slots.
The specifics of how this is done is dependent on the data source used, and will be explained in more detail when the experiments are described in Section \ref{sec:empirical}.

Given the preprocessed location update data containing $N$ users, $T$ time slots, and $C$ location categories, we construct the possible category set for each user at each time slot as
\begin{eqnarray}
\Omega_{ij} \!\!&:=&\!\!  \{k \in [C]~|~\text{category $k$ appears in the location uncertainty circle of user $i$ at time slot $j$}.\}\nonumber
\end{eqnarray}

\subsection{Location Category Inference by Negative-Unlabeled Tensor Factorization}
Given a total of $N$ users, $T$ time slots, $C$ location categories, and the possible location category set $\Omega$, our goal is to infer a location probability tensor ${\bm{\mathcal X}}\in \R^{N\times T \times C}_+$, where each element $\bm{\mathcal X}_{ijk} \in [0,1]$ denotes the probability of user $i$ at time slot $j$ visiting location category $k$. Specifically, the larger the entry $\bm{\mathcal X}_{ijk}$, the greater the chance that user $i$ was visiting the location category $k$ during the time slot $j$. To effectively infer the tensor ${\bm{\mathcal X}}$, we need to restrict $\bmX$ from several key observations.

The first observation is that if $k\notin \Omega_{ij}$, $\bm{\mathcal X}_{ijk}$ must be $0$. This is because a user cannot be visiting a location category if no venue in such category is within the error allowance of the location update. In addition, our second observation is that among the location categories in $\Omega_{ij}$, there is one and only one true category since a user can only appear at one place at a time. In other words, we have $\sum_{k\in\Omega_{ij}} \!\bm{\mathcal X}_{ijk}=1,\ \forall i\in [N],\ j\in [T]$.
As an extreme case, the entry $\bm{\mathcal X}_{ijk}$ equals to $1$ if there is only one location category 
located within the location uncertainty circle. Combining both observations $1$ and $2$, we face a scenario where (i) the positive (i.e., true location category) class contains only one example; (ii) all the labeled examples are sampled from the negative class (i.e., false location categories) and the unlabeled examples come from both negative and positive classes. We refer to such a scenario as \emph{negative-unlabeled} (\emph{NU}) setting. Given the first two observations together with the probability assumption, we have
\begin{eqnarray}\label{eqn:1}
\left\{\begin{array}{l}
\bm{\mathcal X}_{ijk}=0,\ \ \forall i,j,\ \text{and }\ k\notin \Omega_{ij}\\
\label{eqn:reqX}
\sum_{k\in\Omega_{ij}} \bm{\mathcal X}_{ijk}=1,\ \ \forall i, j,\ \text{and }\ \!\!k\in \Omega_{ij} \\
\bm{\mathcal X}_{ijk}\ge 0,\ \ \forall i,j,k.
\end{array}\right.
\end{eqnarray}
We note that the NU setting specified in our paper is related to the problem of partial labeling~\cite{cour2011learning}, where every instance has a candidate set of labels and only one of which is correct. However, our problem is more challenging to solve since the instances in the partial labeling problem have feature data while we have no access to any user features in our modeling.

The observations from the \emph{local} perspective specified in (\ref{eqn:1}) are insufficient to recover the tensor ${\bm{\mathcal X}}$ as we can fill its unobserved entries (i.e., $\bm{\mathcal X}_{ijk},\ k \in \Omega_{ij}$) in with any non-negative values that add up to $1$. To this end, we need to consider the inference of $\bmX$ from a \emph{global} perspective by integrating all users' trajectory data together to collaboratively locate each other, instead of learning a separate model for each user in isolation. 
The collaborative approach captures the fact that people's behaviors typically follow similar patterns based on their lifestyle.
For example, many parents exhibit similar travel patterns caused by picking up or dropping off their children at daycare or school. Such patterns are evident at the category level and we need not consider specific venues.
%
%
%
%
In addition to user-user similarity, time-time similarity is another factor we can take advantage of since users' trajectories are usually consistent from day to day or week to week. 
The observations above suggest that the underlying tensor ${\bm{\mathcal X}}$ should be close to low-rank. To see this, let's consider an ideal case where all the users belong to multiple lifestyle categories and the people with the same lifestyle behave the same. In this case, the rank of the location category tensor is upper bounded by the number of lifestyle categories, a typically small number. Indeed, the low-rank assumption can be verified by another view that people's daily trajectory paths are generally believed to be dictated by a small number of latent factors~\cite{wang2014travel,zheng2015trajectory}.

Combining our observations together, we recover the tensor $\bmX$ by solving the following negative-unlabeled tensor factorization (NUTF) problem:
\begin{eqnarray}\label{eq:main}
\label{eqn:tensor}
\!\!\!&\min \limits_{\bm{\mathcal X},\ \bm{\mathcal Y}\in\R^{N\times T\times C}}& \!\!\!\|\bm{\mathcal X}-\bm{\mathcal Y}\|_F^2\\
\!\!\!&\text{s.t.}& \!\!\!\text{rank}(\bm{\mathcal Y})\leq r\nonumber\\
\!\!\!& &\!\!\!
\left\{\begin{array}{l}
\!\!\!\bm{\mathcal X}_{ijk}=0,\ \forall i,j,\ \text{and }\ k\notin \Omega_{ij}\\
\label{eqn:reqX}
\!\!\!\sum_{k\in\Omega_{ij}} \!\bm{\mathcal X}_{ijk}=1,\ \forall i, j,\ \text{and }\ \!\!k\in \Omega_{ij} \\
\!\!\!\bm{\mathcal X}_{ijk}\ge 0,\ \forall i,j,k,
\end{array}\right.
\end{eqnarray}
where we require $\bmX$ that satisfies the NU constraints \eqref{eqn:reqX} and also close to a low rank tensor $\bm{\mathcal{Y}}$ with rank no more than $r$. Given the recovered tensor $\bm{\mathcal Y}$, we can not only locate users when their location update data is available, but also infer their location categories even when there is no location update data available. Note that location update data is usually very sparse, i.e., a majority of mobile users have only 3.0 -- 10.6  location updates per day, hence there are no location updates for most of time slots. Therefore, the capability of locating users without location update data makes our method extremely appealing.

\subsection{A Parameter-free and Scalable Optimization Algorithm}
In order to efficiently solve the NUTF model \eqref{eq:main}, we adopt an alternating minimization scheme that iteratively fixes one of $\bm{\mathcal X}$ and $\bm{\mathcal Y}$ and minimizes with respect to the other. One nice property is that the proposed algorithm is \emph{optimization parameter free}, that is, the user does not need to decide any optimization parameter such as step length or learning rate.

\begin{algorithm}[t]
\caption{Projection of a vector onto the probability simplex~\cite{DBLP:journals/corr/WangC13a}}
\begin{algorithmic}[1]
\STATE {\bfseries Input:} a vector $\v\in \R^d$ to be projected\\
\STATE Sort $\v$ into $\tv$: $\tv_1\ge \tv_2\ge\cdots\ge \tv_d$
\STATE Find $k=\max\{j\in[d]: \tv_j-\frac{1}{j}(\sum_{i=1}^k \tv_i-1)>0\}$
\STATE Compute $\theta=\frac{1}{k}(\sum_{i=1}^k \tv_i-1)$
\STATE {\bfseries Return:} $\u\in \R^d\ \text{s.t.}\ \u_i=\max(\v_i-\theta,\ 0),\ i\in [d]$
\end{algorithmic} \label{alg:1}
\end{algorithm}
\vspace{3mm}
{\noindent \bf Update $\bmX$} \vspace{1mm}\\
In each iteration of the alternating minimization algorithm, we first update $\bm{\mathcal Y}$ with a fixed $\bm{\mathcal X}$, and then update $\bm{\mathcal X}$ by fixing $\bm{\mathcal Y}$. When $\bm{\mathcal Y}$ is fixed, our goal becomes learning a closest tensor that satisfies the NU constraints \eqref{eqn:reqX}. To this end, we rewrite the objective function \eqref{eqn:tensor} by treating the entries lying within and outside of the $\Omega$ set separately, i.e.,
\begin{eqnarray}
\label{eqn:tensor2}
\!\!\!\!&\!\!\!\!\!\!\!\min \limits_{\bm{\mathcal X}\in\R^{N\times T\times C}}&\!\!\!\!\!\!\!\!\!\!\sum_{i,j;\ k\notin\Omega_{ij}}\!\!\!\!\!(\bm{\mathcal X}_{ijk}-\bm{\mathcal Y}_{ijk})^2+\!\!\!\!\!\!\sum_{i,j;\ k\in \Omega_{ij}}\!\!\!\!\!(\bm{\mathcal X}_{ijk}-\bm{\mathcal Y}_{ijk})^2\\
&&\!\!\!\!
\left\{\begin{array}{l}
\bm{\mathcal X}_{ijk}=0,\ \forall i,j,\ \text{and }\ k\notin \Omega_{ij}\\
\label{eq:matrix}
\sum_{k\in\Omega_{ij}} \!\bm{\mathcal X}_{ijk}=1,\ \forall i, j,\ \text{and}\ k\in \Omega_{ij} \\
\bm{\mathcal X}_{ijk}\ge 0,\ \forall i,j,k
\end{array}\right.
\end{eqnarray}
The optimization problem \eqref{eqn:tensor2} consists of two independent and easily-computable subproblems. For the first subproblem that only involves the entries outside of the possible set $\Omega$, we simply set all of them as zeros to meet the NU constraints \eqref{eqn:reqX}. The second subproblem only involves the entries lying within the possible set $\Omega$ and is essentially a least square problem under a probability simplex constraint. Specifically, for each location update with user $i$ and time slot $j$, we project a $|\Omega_{ij}|$-dimensional vector $\bm{\mathcal X}_{ij\Omega_{ij}}$ onto the probability simplex, which can be efficiently computed in $O(|\Omega_{ij}| \log |\Omega_{ij}|)$ time, as described in Algorithm \ref{alg:1}.

\vspace{3mm}
{\noindent \bf Update $\bm{\mathcal{Y}}$}\vspace{1mm}\\
When $\bmX$ is fixed, we update $\bmY$ by solving the following optimization problem:
\begin{eqnarray}
\label{eqn:matrix}
&\min \limits_{\bmY\in\R^{N\times T\times C}}&\|\bmX-\bmY\|_F^2,\\
&\text{s.t.}& \text{rank}(\bmY)\leq r.\nonumber
\end{eqnarray}
There are multiple ways to define the rank of a tensor, such as by using the Candecomp/Parafac (CP) decomposition~\cite{kolda2009tensor}
\begin{eqnarray}
\text{rank}(\bmY) = \min \left\{r~|~\bm{\mathcal Y} = \sum_{i=1}^r \a_i\circ \b_i\circ \c_i \right\}, \label{eq:cprank}
\end{eqnarray}
where $\a_i\in\R^N$, $\b_i\in\R^T$, $\c_i\in\R^C$, and the symbol $\circ$ represents the vector outer product. However, the problem of computing the rank of a tensor is NP-hard in general
~\cite{hillar2013most,shitov2016hard}.
Although many heuristic algorithms have been developed to increase the efficiency of CP decomposition~\cite{espig2012regularized,phan2013low,rajih2008enhanced,DBLP:conf/nips/0002O14}, they are still not scalable enough to handle our real-world location category inference problem that involves a large number of users and time slots. As a concrete example, recovering a rank $10$ tensor of size $500\times 500\times 500$ takes the state-of-the-art tensor factorization algorithm TenALS\footnote{\scriptsize{\url{http://web.engr.illinois.edu/~swoh/software/optspace/code.html}}} more than $20,000$ seconds on an Intel Xeon $2.40$ GHz processor with $64$ GB main memory. 

\begin{algorithm}[t]
\caption{Efficient Algorithm for Computing the Sparse Low-rank Approximation of the Matrix $\X$}
\label{alg:pm}
\begin{algorithmic}[1]
\STATE {\bfseries Input:} $\X\in\R^{N\times TC}$, support set $\Omega$, rank $r$, number of iterations $T$
\STATE {\bfseries Initialization:}
Gaussian random matrix $\textbf{R}\in \R^{TC\times r}$ satisfying $\textbf{R}_{ij}\sim \mathcal{N}(0,1)$, $\Y \leftarrow \textbf{0}^{N \times TC}$
\STATE
$\B\leftarrow \X\textbf{R} $
\STATE
$ \textbf{Q}\leftarrow \text{\emph{QR}}(\B) $
\FOR{$t=1,2,\cdots, m$}
\STATE
$\B\leftarrow \X(\X^\top \textbf{Q}) $
\STATE
$ \textbf{Q}\leftarrow \text{\emph{QR}}(\B) $
\ENDFOR
\STATE
$\textbf{C}\leftarrow \textbf{Q}^\top\X $
\STATE
$ \Y\in\R^{N\times TC}$:
$\Y_{ik}\leftarrow \textbf{Q}_{i:}\textbf{C}_{:k},\ \exists j,\ k\in \Omega_{ij} $
\STATE {\bfseries Output:} sparse low-rank approximation $\Y$
\end{algorithmic}
\end{algorithm}

In order to significantly improve the scalability of the proposed model in \eqref{eqn:matrix}, we use the rank of the unfolding matrix as the rank of the tensor.  
To this end, we define the unfolding of our 3-order tensor $\bmY\in \R^{N\times T\times C}$ by merging the second (time) and third (location categories) indices of tensors as the column index of matrices. In other words, the matrix $\Y$ is the concatenate of $\bm{\mathcal Y}$'s lateral slices along the time mode, i.e.,
\[
\Y = \text{Unfold}(\bmY) := [\bm{\mathcal Y_{:1:}}\ \cdots\  \bm{\mathcal Y_{:T:}}] \in \R^{N\times TC}.
\]
In this way, the tensor rank in \eqref{eqn:matrix} is defined by the matrix rank
\[
\text{rank}(\bmY) = \text{rank}(\text{Unfold}(\bmY))=\text{rank}(\Y).
\]
Therefore, we can cast the target problem \eqref{eqn:matrix} into an equivalent matrix optimization problem as
\begin{eqnarray}
\label{eqn:matrix2}
&\min \limits_{\Y\in\R^{N\times TC}}&\|\X-\Y\|_F^2,\\
&\text{s.t.}& \text{rank}(\Y) \leq r,\nonumber
\end{eqnarray}
where $\X = \text{Unfold}(\bmX)$. 
Problem \eqref{eqn:matrix2} essentially aims to find matrix $\X$'s best rank-$r$ approximation. Given $\X$'s singular value decomposition (SVD) $\U\bm\Sigma \V^\top$, it is well known that its best rank-$r$ approximation is given by $\U_r\bm\Sigma_r \V_r^\top$, where $\U_r$, $\V_r$ contain the first $r$ columns of $\U$ and $\V$, and $\bm\Sigma_r$ is a diagonal matrix with the first $r$ singular values lying on the diagonal. The SVD step can be calculated in a polynomial time, thus is a significant improvement over the NP-hard problem of computing the tensor rank. However, computing the exact SVD of an $N\times TC$ matrix still takes $O(\max(N,TC)\min(N,TC)^2)$ time, which is far from scalable. 
To this end, we propose to significantly improve its efficiency using the following tricks: (i) since we do not need to compute exact SVD in practice, we instead compute its approximatation using the power method \cite{halko2011finding}, and (ii) we note that in the next step when $\Y$ is fixed, only its entries inside the support $\Omega$ are involved to update the matrix $\X$. In this sense, we do not need to compute $\Y$'s entries outside the support $\Omega$ in the current step, thus allows us to further improve the efficiency. Algorithm \ref{alg:pm} shows the detailed steps of the sparse low-rank approximation algorithm, where the notations $\textbf{Q}_{i:}$ and $\textbf{C}_{:k}$ represent the $i$-th row of $\textbf{Q}$ and the $k$-th row of $\textbf{C}$, respectively. $\text{\emph{QR}}(\cdot)$ indicates the reduced QR factorization. In Algorithm \ref{alg:pm}, we assume that $N\ge TC$ by default, since in real-world applications the number of users $N$ is usually a dominant factor. When $N< TC$, we will use $\X^\top$ as the input to Algorithm \ref{alg:pm} such that the time complexity is dependent to $\min(N, TC)$.

\vspace{3mm}
{\noindent \bf Total Complexity}\vspace{1mm}\\
Since $\X$ is a sparse matrix with only $|\Omega|$ non-zero entries, the computational cost of the steps 3, 6, 9, and 10 in Algorithm~\ref{alg:pm} are merely $O(|\Omega|r)$. In addition, since $\B\in \R^{N\times r}$ is a tall-and-skinny matrix, its QR factorization can be efficiently computed using $O(Nr^2)$ operations. Indeed, we can further reduce this cost if $N< TC$, in which case we use $\X^\top$ instead of $\X$ as Algorithm \ref{alg:pm}'s input. In this way, the QR decomposition can be computed using $O(TCr^2)$ operations. Combining all the computations together, we can update $\Y$ within $O(|\Omega|rm+\min(N, TC)r^2m)$ time, where $m$ is the number of iterations. Since $|\Omega|\ll NTC$ and $r\ll \min(N,TC)$, the proposed algorithm is significantly faster than the naive SVD computation with $O(\max(N,TC)\min(N,TC)^2)$ complexity.

Given $\Y$, we update $\X$ by projecting $\Y$ to a space that satisfies the NU constraints \eqref{eq:matrix}. Although this step is efficient enough in the tensor case, it becomes even simpler in the matrix setting. Since $\Y$ is already a sparse matrix with only $|\Omega|$ non-zero entries, we only need to project each vector within $\Omega_{ij}, \forall i,j$ onto the probability simplex. The time complexity of this step is $O(\sum_{i,j}|\Omega_{ij}|\log (|\Omega_{ij}|))$. Since we have $|\Omega_{ij}|\le C\ \forall i, j$, this time complexity is upper bounded by $O(|\Omega|\log C)$.

Overall, each iteration (updating $\bmX$ and $\bmY$ once) of the proposed alternating minimization algorithm can be efficiently computed within
\[O(|\Omega|(rm+\log C) + \min(N, TC)r^2m)\]
time. In addition to the low computational cost in each iteration, our algorithm converges very fast as well, as verified by extensive experiments on both simulated and real-world data sets. For example, it takes only $3$ iterations to optimize a problem with $3$ million users, $500$ time slots, $200$ location categories, and more than a billion non-zero entries in $\Omega$, with a running time for each single iteration about $18$ minutes.

\section{Experiments}\label{sec:empirical}
In this section, we first conduct a simulated study to verify that the proposed algorithm is scalable to large-scale location category inference problems and also robust to noise.
We then evaluate the proposed algorithm on multiple real-world mobility data sets. All the experiments were run on a Linux server with an Intel Xeon 2.40 GHz CPU and 64 GB of main memory.

\subsection{Experiments with Synthesized Data}
 \begin{table*}[t]
   \caption{CPU time and prediction accuracies with different number of users and location updates. K, M, B indicates thousands, millions, and billions, respectively. \label{tab:scalability}}
    \centering
  \begin{tabular}{|l|c|c|c|c|c|}
      \hline
     \#users  &  \#time slots  & \#location categories &  $|\Omega|$ &   CPU Time (s) & Prediction Accuracy \\
      \hline
   100 K & 500 & 200 & 40 M& 124&$100\%$\\
   \hline
   200 K & 500 & 200 & 80 M& 240&$100\%$\\
   \hline
   500 K & 500 & 200 & 200 M& $660$& $100\%$\\
   \hline
   1 M & 500 & 200 & 400 M & $1,310$& $100\%$\\
   \hline
   2 M & 500 & 200 & 800 M & $1,938$& $100\%$\\
   \hline
   3 M & 500 & 200 & 1.2 B & $3,397$& $100\%$\\
   \hline
  \end{tabular}
\end{table*}
We first conduct experiments with simulated data to verify that the proposed
location category inference algorithm is computationally efficient and robust to location errors. To this end, we fix $T$ and $C$, the number of time slots and the number of location categories, to 500 and 200, respectively. We also vary the number of mobile users, $N$, in range $\{100\ \text{K},\ 200\ \text{K},\ 500\ \text{K},\ 1\ \text{M},\ 2\ \text{M},\ 3\ \text{M}\}$, where $\text{K}$ and $\text{M}$ stand for thousand(s) and million(s), respectively. For a fixed $N$, we randomly assign all the mobile users to $10$ lifestyle classes, with the class memberships blind to our algorithm. We assume that users in the same lifestyle class visit the same location categories at the same time while users in different classes behave differently. For each user and $20\%$ of the randomly sampled time slots, we generate her noisy location updates that contain $4$ candidate location categories for each of them. Among the $4$ candidate categories, one is the true location category and the other three are randomly sampled from the remained $C-1$ location categories. We input the generated noisy location updates data to our algorithm and compare the predicted results with the ground truth information. Table~\ref{tab:scalability} summarizes the CPU time and prediction accuracies of inferring location categories on this data. Specifically, the prediction accuracy is defined as
\[\frac{\text{\#(category with the highest prob. = true category)}}{\text{\#location updates}}\times 100\%.\]

Table~\ref{tab:scalability} clearly shows that the proposed algorithm can \emph{perfectly} recover the underlying true location categories as all the prediction accuracies equal to $100\%$. Besides, the proposed algorithm is extremely efficient, e.g., even with $3$ million users and more than $1$ billion candidate location categories, it only takes the proposed algorithm less than $1$ hour to infer the perfect location category in a single thread.

\subsection{Experiment with Real-World Location Update Data}

\vspace{0.3cm}
{\noindent \bf \large{Data Preprocessing}}\vspace{0.1cm}\\
We then conduct experiments with real-world location update data. The raw location update data consists of a temporal stream of records in the form \{\emph{anonymous user id, UTC timestamp, estimated location coordinates, location error}\}. In order to apply our location category inference algorithm, 
the raw data was preprocessed, encompassing three key steps: 1) removal of noise to obtain meaningful location coordinates, 2) association of  meaningful locations with candidate venues (and hence categories), and 3) determination of the local time and quantization into time slots.

The first step is necessary because we are only interested in locations where the user has spent significant time -- locations that could have actually been visited in a meaningful way. In addition, the dynamics of mobile phone location update algorithms may not be completely understood.  For example, the significant-change location service, a frequently used location API on iOS devices, ``delivers updates only when there has been a significant change in the device's location, such as 500 meters or more.''\footnote{\url{https://developer.apple.com/library/content/documentation/UserExperience/Conceptual/LocationAwarenessPG/CoreLocation/CoreLocation.html}} Also, under this set of location APIs, the precise triggers which initiate an update are not fully disclosed.  
To determine meaningful locations, the first step is to estimate the dwell time by taking the difference of subsequent timestamps. Once dwell time is estimated, locations where the dwell time is less than a threshold are removed.

The next step in the process is to associate the location update coordinates to venues. In our case, Foursquare APIs\footnote{\url{https://developer.foursquare.com/docs/}} were used to query for potential venues. Specifically, we define the location uncertainty circle for an update as the circle centered on the update coordinates with radius given by the reported location error, $r_{\text{error}}$. Similarly, we represent a venue as a circle with a fixed radius $r_{\text{venue}}$ centered on the venue coordinates. Then, a venue is considered a candidate venue if it intersects with the location uncertainty circle, i.e., if $h(\x, \mathbf{p}) \le r_{\text{error}} + r_{\text{venue}}$. Here, $h(\cdot,\cdot)$ is the haversine distance between the location update $\x$ and the venue coordinates $\mathbf{p}$. Once a candidate set of venues is known for a location update, the category information is extracted. We used a subset of 42 of the available categories from the Foursquare hierarchy. The categories were chosen to cover the entire hierarchy, and any categories falling below a chosen category were mapped to its closest ancestor category. Categories chosen include: Museum, College \& University, Music Venue, Train Station, Theater, Zoo, Library, Post Office, etc.

Finally, we convert all timestamps into local time to better understand the context of a user's visit. This is done by determining the local timezone based on the coordinates and then adjusting the UTC timestamp using the appropriate offset. Finally the timestamps are quantized, non-uniformly, into time slots across each day using the following process. The time period from 1am to 7am is mapped to the first bin, 7am-9am is mapped to the second bin, 9am-11am is mapped to the third bin, etc., giving a total of 10 bins per day. The non-uniform scheme is chosen since there is little activity during the early morning.

\vspace{3mm}
{\noindent \bf \large{Experiments}}\vspace{1mm}\\
\begin{table}[t]
\caption{Statistics of the three location updates data sets}\label{Statistics}
\begin{center}
\begin{tabular}{|l|c|c|c|}
    \hline
{\!\!Data sets} &\!\!\#users\!\!  &  \!\#time slots\!\!  &  \!\#loc. categories\!\! \\
\hline
\hline
\!\!New York, NY  &$77,084$ & $138$ &$42$ \\
\hline
\!\!Austin, TX  &$10,211$ & $137$ & $42$ \\
\hline
\!\!San Francisco, CA\!\! &$11,083$ & $138$ & $42$ \\
\hline
\end{tabular}
\end{center}
\end{table}

\begin{table}[t]
\caption{\label{table:perf_real_world}Prediction accuracies of the proposed algorithm NUTF and three baseline algorithms on the three real-world location updates data. N/A indicates that the factorization task could not be completed due to memory limitations.}
\begin{center}\label{acc}
\begin{tabular}{|l|c|c|c|c|c|}
\hline
{Data sets} &k&\!\! NUTF\!\! & \!\!CP-APR\!\! &\!\! Rubik\!\! &\!\! BPTF\!\!\\
\hline
\hline
\multicolumn{1}{|l|}{\multirow {5}{*}{\!\!New York, NY\!\!}}&1 &$\bm {35\%}$&N/A&$ 12\%$&$ 10\%$\\
\cline{2-6}
&2 &$\bm {54\%}$&N/A&$15\%$&$ 14\%$\\
\cline{2-6}
&3 &$\bm {67\%}$&N/A&$19\%$&$ 17\%$\\
\cline{2-6}
&4 &$\bm {74\%}$&N/A&$21\%$&$ 19\%$\\
\cline{2-6}
&5 &$\bm {79\%}$&N/A&$24\%$&$ 21\%$\\
\hline
\hline
\multicolumn{1}{|l|}{\multirow {5}{*}{\!\!Austin, TX\!\!}}&1 &$\bm {34\%}$&$11\%$&$12\%$&$9\%$\\
\cline{2-6}
&2 &$\bm {52\%}$&$13\%$&$14\%$&$11\%$\\
\cline{2-6}
&3 &$\bm {65\%}$&$16\%$&$18\%$&$16\%$\\
\cline{2-6}
&4 &$\bm {73\%}$&$20\%$&$21\%$&$19\%$\\
\cline{2-6}
&5 &$\bm {78\%}$&$23\%$&$23\%$&$22\%$\\
\hline
\hline
\multicolumn{1}{|l|}{\multirow {5}{*}{\!\!San Francisco, CA\!\!}}&1 &$\bm {32\%}$&$11\%$&$11\%$&$9\%$\\
\cline{2-6}
&2 &$\bm {50\%}$&$12\%$&$14\%$&$11\%$\\
\cline{2-6}
&3 &$\bm {63\%}$&$15\%$&$17\%$&$17\%$\\
\cline{2-6}
&4 &$\bm {70\%}$&$20\%$&$21\%$&$19\%$\\
\cline{2-6}
&5 &$\bm {75\%}$&$22\%$&$23\%$&$21\%$\\
\hline
\end{tabular}
\end{center}
\end{table}
We run experiments across three major U.S.\ cities, New York, NY, Austin, TX, and San Francisco, CA, using anonymized mobile location data that was collected as part of our partnership with a leading U.S.\ location-based services company. 
Since there is no ground truth information (i.e., true location categories visited by mobile users) available, we evaluate the prediction performance by 
using the certain location updates that contain only one location category within their corresponding location uncertainty circles. These true location categories are used as the validation set. For the same user-timeslot pairs in the validation set, we also create noisy data by marking all the $C$ location categories as the possible categories. We combine this data set with all the noisy location updates that have multiple location categories within the location uncertainty circles, and use the combined data as the input to our algorithm. After learning the tensor $\bm{\mathcal X}$, we check, for each location update in the validation set, if the categories with the top $k$ highest probabilities contain the true venue category. We set $k=\{1,2,\cdots,5\}$ in our study. In particular, $k=1$ means that the true venue category is consistent with the learned highest probability category.

Since our proposed NUTF is, to the best of our knowledge, the first algorithm that can infer location categories purely based on inaccurate mobility location data, there is no direct baseline for comparison. Note that our validation set contains the location updates with only one location category within their corresponding location uncertainty circles. We cannot compare our method with 1-NN approach since its results are trivial, i.e., the only one location category in a location uncertainty circle is always the nearest neighbor. In this case, we compare our negative-unlabeled tensor factorization approach with three state-of-the-art tensor factorization algorithms: (a) \textbf{CP-APR}, Candecomp-Parafac alternating Poisson regression \cite{DBLP:journals/siammax/ChiK12}, (b) \textbf{Rubik}, knowledge-guided tensor factorization and completion method \cite{DBLP:conf/kdd/WangCGDKCMS15}, and (c) \textbf{BPTF}, Bayesian probabilistic tensor factorization \cite{DBLP:conf/sdm/XiongCHSC10}. To this end, given a set of $N$ users, $T$ time slots, and $C$ location categories, we generate a $N\times T\times C$ partially-observed tensor $\bm{\mathcal W}$ as
\begin{eqnarray}\label{eqn:8}
        \bm{\mathcal W}_{i jk} \!\!=\!\! \begin{cases}
          0.1, & \!\!\text{if}\ k \in \Omega_{ij}, \text{and with a probability}\ p\\
          \text{unobserved}, \!\!& \!\!\text{if}\ k \in \Omega_{ij}, \text{and with a probability}\ 1-p\nonumber\\
          0, & \!\!\text{if}\ k \notin \Omega_{ij}.
        \end{cases}
\end{eqnarray}
We use this tensor to simulate our condition where all the location categories not covered by the location uncertainty circle are impossible (= 0), and the location categories covered by the circle have unknown probabilities with some samples having the same initialization (= 0.1). We then use the tensor $\bm{\mathcal W}$, with all the entries in the validation set marked as unobserved, as an input to our baseline tensor factorization algorithms to recover its low-rank approximations. To properly select the parameter $p$ in (\ref{eqn:8}), we examine two extreme cases. When $p=0$, we only observe zero entries in the tensor $\bm{\mathcal W}$ and factorizing it yields an all-zero tensor, which is a trivial solution. On the other hand, when $p=1$, the tensor $\bm{\mathcal W}$ reduces to a fully-observed tensor which should be far away from the underlying ideal tensor that has only one non-zero element in each $\Omega_{ij}$. In order to select the optimal $p$, we perform a $5$-fold cross validation over the ranges $p\in \{0.05,\ 0.1, \ldots,\ 0.95\}$.

Given the tensor recovered by the baseline algorithms, we check, for each location update in the validation set, if the categories with the top $k$ highest probabilities contain the true venue category --- identical to the way used to evaluate our method. In this experiment, we set 10 time slots per day where the first slot contains the time period from 1 am to 7 am, and every other time slot contains two hours. We consider location updates which corresponded to a dwell time of at least 20 minutes. If there are more than one location updates within a same time slot, we select the one with the longest dwell time. For each city, we look at $2$ weeks of location update data and a random sample of users, with the statistics of the three data sets summarized in Table~\ref{Statistics}. We only get a total of 138 or 137 time slots for the three cities due to lacking of data at the ends. Notably, our proposed algorithm is scalable enough to handle problems with large numbers of users, time slots, and location categories. The two week period and relatively small sample size were chosen to result in data volumes which were compatible with baseline models that don't scale as well, such as CP-APR and BPTF.
In our experiments, we set rank $r=20$ and number of iterations $T=100$ for all the data sets. The parameters of all the baselines are tuned to give the best performance.

Table \ref{table:perf_real_world} summarizes the prediction performance averaged over the five trials of all the algorithms. We first observe that the prediction accuracies of the proposed method NUTF are very encouraging. Among all the three data sets, its exact matching ($k$=1) accuracies range from $32\%$ to $35\%$, and its top $5$ category accuracies range from $75\%$ to $79\%$. Given that these data sets have a total of $42$ location categories and a random guess only yields a $2.4\%$ accuracy, our prediction performance clearly demonstrates the effectiveness of the proposed method. In addition, our method NUTF also yields significantly better performance than all the baseline algorithms. In particular, its accuracies are about $3$ times better than the baselines. Finally, we observe that the prediction performance of all the algorithms is fairly consistent across the three cities, suggesting that our collected data is sufficiently representative.

\subsection{Experiments with Real-World Check-in Data}
\begin{table}[t]
\caption{Average prediction accuracies of the proposed algorithm NUTF, and the baseline algorithms PU-MC and WR-MF when $\textbf{k}$ = 1, 2, $\cdots$, 5.}
\begin{center}\label{acc}
\begin{tabular}{|l|c|c|c|c|}
\hline
{Data sets} &\ k\ & \ NUTF \ & \ PU-MC\  & \ WR-MF \ \\
\hline
\hline
\multicolumn{1}{|l|}{\multirow {5}{*}{NYC Check-ins}}&1 &$\bm {30\%}$&$8\%$&$19\%$\\
\cline{2-5}
&2 &$\bm {43\%}$&$14\%$&$30\%$\\
\cline{2-5}
&3 &$\bm {50\%}$&$18\%$&$37\%$\\
\cline{2-5}
&4 &$\bm {55\%}$&$21\%$&$42\%$\\
\cline{2-5}
&5 &$\bm {58\%}$&$25\%$&$46\%$\\
\hline
\hline
\multicolumn{1}{|l|}{\multirow {5}{*}{Tokyo Check-ins}}&1 &$\bm {46\%}$&$12\%$&$40\%$\\
\cline{2-5}
&2 &$\bm {60\%}$&$19\%$&$53\%$\\
\cline{2-5}
&3 &$\bm {67\%}$&$23\%$&$62\%$\\
\cline{2-5}
&4 &$\bm {71\%}$&$27\%$&$67\%$\\
\cline{2-5}
&5 &$\bm {74\%}$&$29\%$&$71\%$\\
\hline
\end{tabular}
\end{center}
\end{table}
In addition to the experiments with the mobility location update data, we conduct experiments on two real-world check-in data sets to further verify that the proposed algorithm can also reliably infer users' location categories even when there is no location update. The two data sets collected by~\cite{yang2015modeling} consist of the Foursquare check-ins in New York City (NYC) and Tokyo from $12$ April $2012$ to $16$ February $2013$. The New York check-in data set contains 824 users, $38,336$ venues and $227,428$ check-ins, while the Tokyo check-in data contains $1,939$ users, $61,858$ venues, and $573,703$ check-in records. Each check-in includes a user ID, a time stamp, a venue ID, and the category of the venue. By manually merging the similar and infrequent venue categories together, we finally obtain $122$ venue categories for both data sets. In this experiment, we use hourly time slots and select the check-in with the longest dwell time if there is more than one check-in within the same time slot.

In order to evaluate our performance, we randomly sample $10\%$ of the check-ins as the validation set, and mark all the $C$ location categories as the possible categories for all of them. This mimics the situation when there is no available location update, or the location update is too inaccurate to contain any meaningful location information. We then combine the noisy and accurate check-ins altogether and use them as the input to our algorithm. Similar to our last experiments, after learning the tensor $\bm{\mathcal X}$, we check that, for each check-in in the validation set, if the categories with the top $k$ highest probabilities contain the true venue category. 

\begin{table}[t]
\caption{Average CPU time (in seconds) of the proposed algorithm NUTF, and the baseline algorithms PU-MC and WR-MF}\label{time}
\begin{center}
\begin{tabular}{|l|c|c|c|}
    \hline
{Data sets} &\ \ NUTF\ \ &\ \ \ PU-MC\ \ \ &\ \ \ WR-MF\ \ \ \\
\hline
\hline
NYC Check-ins  &\bf{4.8} & $599$ &14  \\
\hline
Tokyo Check-ins  &{\bf 12} & $498$ & 48 \\
    \hline
\end{tabular}
\end{center}
\end{table}

To find baseline algorithms for this study, we cast this problem into a matrix completion problem~\cite{candes2012exact}. Let $N$, $T$, and $C$ be the number of users, time slots, and location categories, respectively. Given the training check-in records that do not belong to the validation set, we construct a $N\times TC$ partially-observed matrix with the observed entries corresponding to the check-ins. In this sense, the prediction problem becomes a matrix completion problem that aims to recover the unobserved entries based on the observed ones. Also, since all the observed entries are drawn from the positive (interest) class, this is a positive-unlabeled (PU) learning problem. We thus compare our method with the state-of the-art PU matrix completion algorithms: (a) \textbf{PU-MC}, PU learning for matrix completion \cite{DBLP:conf/icml/HsiehND15}, and (b) \textbf{WR-MF}, weighted regularized matrix factorization \cite{hu2008collaborative}. Specifically, we feed the generated partially-observed matrix to these algorithms, and compare the ground-truth venue categories in the validation set with their corresponding outputs. All the experiments in this study are repeated five times, and the prediction accuracies and the CPU time averaged over the five trials are reported in Table~\ref{acc} and Table~\ref{time}, respectively.

Table~\ref{acc} summarizes the average prediction accuracies of the proposed algorithm NUTF, and the baseline algorithms PU-MC and WR-MF when $k=1, 2, \cdots, 5$. It clearly shows that the proposed algorithm NUTF significantly outperforms all the baseline algorithms. Specifically, $30\%$ and $46\%$ of the learned highest probability categories are consistent with the ground truth categories of the NYC check-in and Tokyo check-in data sets, respectively. Considering that we have $122$ categories in total and a random guess only yields a $0.8\%$ accuracy, our achieved accuracies are pretty impressive. In addition, Table~\ref{time} summarizes the average CPU time of all the algorithms evaluated here. Among them, our proposed method NUTF is the most efficient algorithm. In particular, NUTF is able to infer the location categories of the Tokyo check-in data set with $1,939$ users, $61,858$ venues, and $573,703$ check-in records in $12$ seconds.

Another interesting finding in this study is that all the algorithms perform better on the Tokyo data than on the NYC data. This is because the Tokyo data set is larger with more users and more check-in records. Since all the algorithms evaluated here are collaborative approaches that learn a universal model from all the available user data, they usually deliver more accurate results as more data is provided.

\section{Conclusions}\label{sec:co}
In this paper, we study the problem of inferring the user visited location categories purely based on their highly inaccurate mobility location data. To the best of our knowledge, our paper is the first study of this kind. To solve this problem, we propose a novel tensor factorization framework NUTF that is able to infer the most likely location categories within the location uncertainty circle. To this end, we highlight several key observations, including the negative-unlabeled constraints and the correlations among users. In order to efficiently solve the tensor factorization problem, we propose a parameter-free and scalable optimization algorithm by effectively exploring the sparse and low-rank structure of the underlying tensor. Our empirical studies conducted on multiple synthesized and real-world data sets confirm both the effectiveness and efficiency of the proposed algorithm.

\bibliographystyle{plain}
\bibliography{Location,durable,Sigir_Rank}
\end{document}